\documentclass[11pt,a4paper]{article}
\usepackage[hyperref]{naaclhlt2018}
\usepackage{times}
\usepackage{latexsym}
\usepackage{amsfonts}
\usepackage{amsmath}
\usepackage{graphicx}
\usepackage{booktabs}
\usepackage{multirow}

\usepackage{url}

\aclfinalcopy 
\def\aclpaperid{1782} 

\newcommand{\red}[1]{\textcolor{black}{#1}}

\title{Personalized neural language models for real-world query auto completion}

\author{Nicolas Fiorini \\
  National Center for Biotechnology Information \\
  National Library of Medicine, NIH \\
  Bethesda, MD, USA \\
  {\tt nicolas.fiorini@nih.gov} \\\And
  Zhiyong Lu \\
  National Center for Biotechnology Information \\
  National Library of Medicine, NIH \\
  Bethesda, MD, USA \\
  {\tt zhiyong.lu@nih.gov} \\}

\date{}

\begin{document}
\maketitle
\begin{abstract}
Query auto completion (QAC) systems are a standard part of search engines in industry, helping users formulate their query. Such systems update their suggestions after the user types each character, predicting the user's intent using various signals --- one of the most common being popularity. Recently, deep learning approaches have been proposed for the QAC task, to specifically address the main limitation of previous popularity-based methods: the inability to predict unseen queries. In this work we improve previous methods based on neural language modeling, with the goal of building an end-to-end system. We particularly focus on using real-world data by integrating user information for personalized suggestions when possible. We also make use of time information and study how to increase diversity in the suggestions while studying the impact on scalability. Our empirical results demonstrate a marked improvement on two separate datasets over previous best methods in both accuracy and scalability, making a step towards neural query auto-completion in production search engines. 
\end{abstract}

\section{Introduction}
Predicting the next characters or words following a prefix has had multiple uses from helping handicapped people \cite{swiffin1987adaptive} to, more recently, helping search engine users \cite{cai2016survey}. In practice, most search engines today use query auto completion (QAC) systems, consisting of suggesting queries as users type in the search box \cite{fiorini2017}. The task suffers from high dimensionality, because the number of possible solutions increases as the length of the target query increases. Historically, the query prediction task has been addressed by relying on query logs, particularly the popularity of past queries \cite{bar2011context,lu2009finding}. The idea is to rely on the wisdom of the crowd, as popular queries matching a typed prefix are more likely to be the user's intent.

This traditional approach is usually referred to as \emph{MostPopularCompletion} (MPC)\cite{bar2011context}. However, the performance of MPC is skewed: it is very high for popular queries and very low for rare queries. At the extreme, MPC simply cannot predict a query it has never seen. This becomes a bigger problem in academic search \cite{lankinen2016character}, where systems are typically less used, with a wider range of possible queries. Recent advances in deep learning, particularly in semantic modeling \cite{mitra2015query} and neural language modeling \cite{park2017neural} showed promising results for predicting rare queries. In this work, we propose to improve the state-of-the-art approaches in neural QAC by integrating personalization and time sensitivity information as well as addressing current MPC limitations by diversifying the suggestions, thus approaching a production-ready architecture.

\section{Related work}
\subsection{Neural query auto completion}
While QAC has been well studied, the field has recently started to shift towards deep learning-based models, which can be categorized into two main classes: semantic models (using Convolutional Neural Nets, or CNNs) \cite{mitra2015query} and language models (using Recurrent Neural Nets, or RNNs) \cite{park2017neural}. Both approaches are frequently used in natural language processing in general \cite{kim2016character} and tend to capture different features. In this work, we focus on RNNs as they provide a flexible solution to generate text, even when it is not previously seen in the training data.

Yet, recent work in this field \cite{park2017neural} suffers from some limitations. Most importantly, the probability estimates for full queries are directly correlated to the length of the suggestions, consequently favoring shorter queries in some cases and hampering some predictions \cite{park2017neural}. By appending these results to MPC's and re-ranking the list with LambdaMART \cite{burges2010ranknet} in another step as suggested in previous work \cite{mitra2015query}, they achieve state-of-the-art performance in neural query auto completion at the cost of a higher complexity and more computation time.

\subsection{Context information}
Still, these preliminary approaches have yet to integrate standards in QAC, e.g. query personalization \cite{koutrika2005unified,margaris2018query} and time sensitivity \cite{cai2014time}. This integration has to differ from traditional approaches by taking full advantage of neural language modeling. For example, neural language models could be refined to capture interests of some users as well as their actual language or query formulation. The same can apply to time-sensitivity, where the probability of queries might change over time (e.g. for queries such as ``tv guide'', or ``weather''). Furthermore, the feasibility of these approaches in real-world settings has not been demonstrated, even more so on specialized domains. 

By addressing these issues, we make the following contributions in this work compared to the previous approaches: 
\begin{itemize}
\item We propose a more straightforward architecture with \textbf{improved scalability};
\item Our method integrates \textbf{user information} when available as well as \textbf{time-sensitivity};
\item We propose to use a balanced beam search for ensuring \textbf{diversity};
\item We test on \textbf{a second dataset} and compare the generalizability of different methods in a specialized domain;
\item Our method achieves \textbf{stronger performance} than the state of the art on both datasets.  
\end{itemize}
\red{Finally, our source code is made available in a public repository\footnote{\url{https://github.com/ncbi-nlp/NQAC}}. This allows complete \textbf{reproducibility of our results} and future comparisons.}


\section{Methods}

\subsection{Personalized neural Language Model}
The justification of using a neural language model for the task of predicting queries is that it has been proven to perform well to generate text that has never been seen in the training data \cite{sutskever2011generating}. Particularly, character-level models work with a finer granularity. That is, if a given prefix has not been seen in the training data (e.g. a novel or incomplete word), the model can use the information shared across similar prefixes to make a prediction nonetheless.\\ 

{\bf Recurrent Neural Network} The difficulty of predicting queries given a prefix is that the number of candidates explodes as the query becomes longer. RNNs allow to represent each character (or word) of a sequence as a cell state, therefore reducing the dimensionality of the task. However, they also introduce the vanishing gradient problem during backpropagation, preventing them from learning long-term dependencies. Both gated recurrent units (GRU) \cite{cho2014properties} and long-short term memory cells (LSTMs) solve this limitation --- albeit with a different approach --- and are increasingly used. In preliminary experiments, we tried various forms of RNNs: vanilla RNNs, GRUs and LSTMs. GRUs performed similarly to LSTM with a smaller computational complexity due to fewer parameters to learn as was previously observed \cite{jozefowicz2015empirical}.\\

{\bf Word embedded character-level Neural Language Model} The main novelty in \cite{park2017neural} is to combine a character-level neural language model with a word-embedded space character. The incentive is that character-level neural language models benefit from a finer granularity for predictions but they lack the semantic understanding words-level models provide, and vice versa. Therefore, they encode text sequences using one-hot encoding of characters, character embedding and pre-trained word embedding (using word2vec \cite{mikolov2013distributed}) of the previous word when a space character is encountered. Our preliminary results showed that the character embedding does not bring much to the learning, so we traded it with the context feature vectors below to save some computation time while enriching the model with additional, diverse information.\\

{\bf User representation} We make the assumption that the way a user types a query is a function of their actual language/vocabulary, but also a function of their interests. Therefore, a language model could capture these user characteristics to better predict the query, if we feed the learner with the information. Each query $q_i$ is a set of words such that $q_i = \{w_1, ..., w_n\}$. $U$ is a column matrix and a user $u \in U$ is characterized by the union of words in their $k$ past queries, i.e. $Q^u = \cup^k_{i=1}{q_i}$. The objective is to reduce, for each user, the vocabulary used in their queries to a vector of a dimensionality $d$ of choice, or $Q^u \rightarrow \mathbb{R}^{d}$. We chose $d=30$, in order to stay in the same computation order of previous work using character embedding \cite{park2017neural}. To this end, we adapted the approach \emph{PV-DBOW} detailed in \cite{le2014distributed}. That is, at each training iteration, a random word $w_i$ is sampled from $Q^u$. The model is trained by maximizing the probability of predicting the user $u$ given the word $w_i$, i.e.:
\begin{equation}
\frac{1}{|U|}\sum_{u \in U}{\sum_{w_i \in Q^u}{log~P(u|w_i )}}.
\end{equation}
The resulting vectors are stored for each user ID and are used as input for the neural net (NN) (see Architecture section).\\

{\bf Time representation} As an example, in the background data (see Section \ref{sec:dataset}), the query ``tv guide'' appears 1,682 times and it is vastly represented in evening and nights. For this reason, we propose to integrate time features in the language model. While there has been more elaborated approaches to model it in the past \cite{shokouhi2012time}, we instead propose a straightforward encoding and leave the rest of the work to the neural net. For each query, we look at the time it was issued, consisting of hour $x$ , minute $y$ and second $z$, and we derive the following features:
\begin{equation}
\begin{gathered}
sin\left(\frac{2\pi(3600x + 60y + z)}{86400}\right),\\
cos\left(\frac{2\pi(3600x + 60y + z)}{86400}\right).
\end{gathered}
\end{equation}
This encoding has the benefit of belonging to $[-1,1]$, which is a range comparable to the rest of the features. It is also capable to model cyclic data, which is important particularly around boundaries (e.g. considering a query at 11:55PM and another at 00:05AM). We proceed the same way to encode weekdays and we end up with four time features.\\

\begin{figure*}
\includegraphics[width=6.3in]{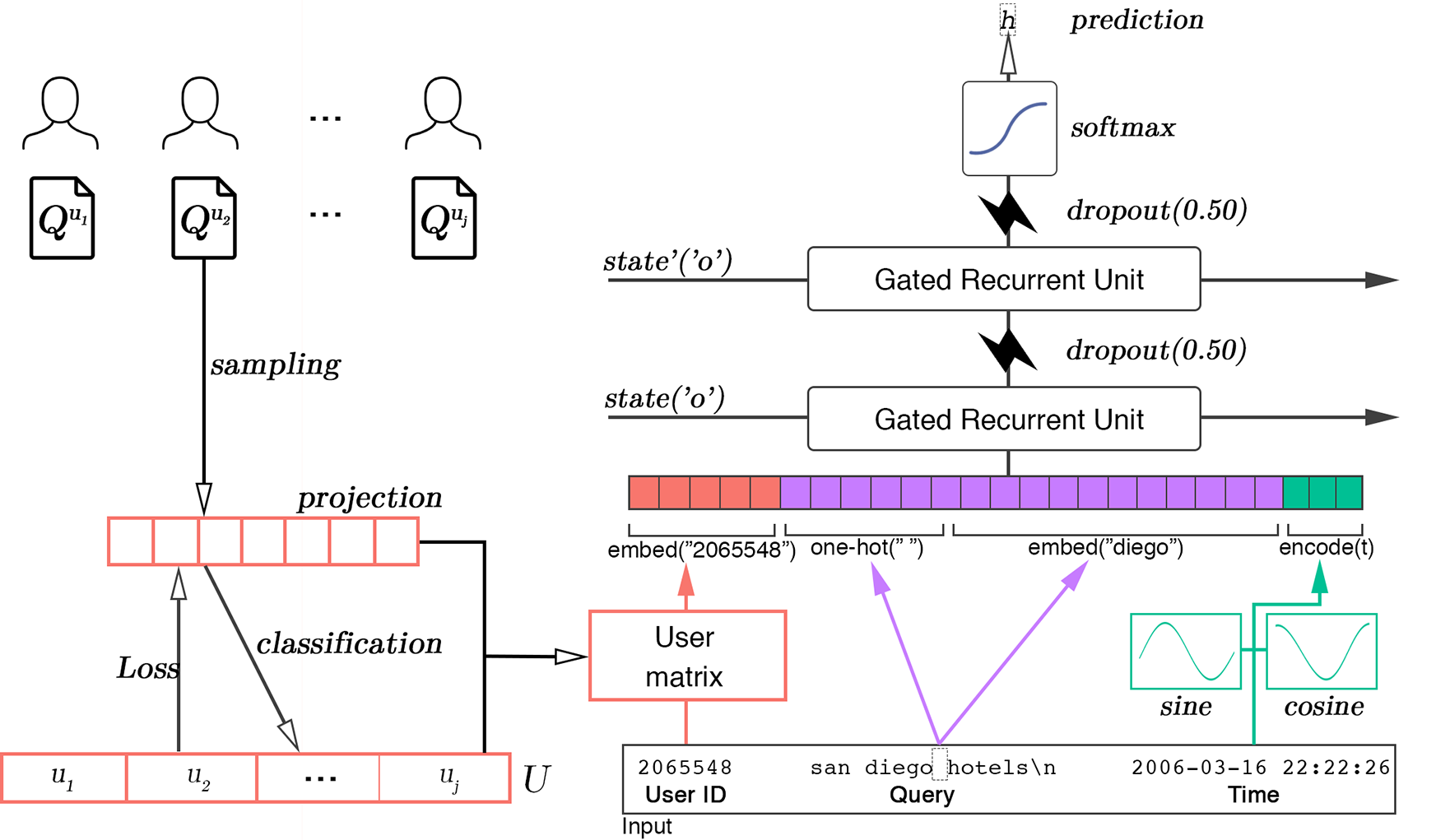}
\caption{Architecture of our proposed model.}
\label{fig:nqac}
\end{figure*}

{\bf Overall architecture} An overview of the architecture is proposed in Figure \ref{fig:nqac}. The input of our neural language model is a concatenation of the vectors defined above, for each character and for each query in the training set. We use zero-padding after the ``\textbackslash n'' character to keep the sequence length consistent, and the NN learns to recognize it. We feed this input vector into 2 layers of 1024 GRUs\footnote{It was reported that using more cells may not help the prediction while hurting computation \cite{park2017neural}.}, each followed by a dropout layer (with a dropout rate of 50\%) to prevent overfitting. Each GRU cell is activated with $ReLu(x) = x^+$ and gradients are clipped to a norm of $0.5$ to avoid gradient exploding problems. The output of the second dropout layer is fed to a temporal softmax layer, which allows to make predictions at each state. The softmax function returns the probability $P(c_i|c_1,...,c_{i-1})$ of the character $c_i$ given the previous characters of the sequence, which is then used to calculate the loss function by comparing it to the next character in the target query. Instead of using the objective denoted in \cite{park2017neural}, we minimize the loss $\mathcal{L}$ defined as the average cross entropy of this probability with the reference probability $\hat{P}(c_i)$ across all queries, that is
\begin{multline}
\mathcal{L} = \\
-\frac{1}{|Q|} \sum_{q \in Q}{\sum_{i = 1}^{|q|-1}{\hat{P}(c_{i+1}) \times log~ P(c_{i+1}|c_1,...,c_i)}}.
\end{multline}
$Q$ is the set of queries in the training dataset, \red{$|Q|$ is the total number of queries in the set} and $|q|$ is the number of characters in the query $q$. Convergence stabilizes around 5-10 epochs for the AOL dataset (depending on the model) and 15-20 epochs for the biomedical specialized dataset (see Section \ref{sec:dataset}).

\subsection{Balanced diverse beam search}
The straightforward approach for decoding the most likely output sequence --- in this case, a suffix given a prefix --- is to use a greedy approach. That is, we feed the prefix into the trained NN and pick the most likely output at every step, until the sequence is complete. This approach has a high chance to output a locally optimal sequence and a common alternative is to use a beam search instead. We propose to improve the beam search by adding a greedy heuristic within it, in order to account for the diversity in the results. A similar suggestion has been made in \cite{vijayakumar2016diverse}, and our proposition differs by rebalancing the probabilities after diversity was introduced. In \cite{vijayakumar2016diverse}, at every step the most likely prediction is not weighted while all others are, by greedily comparing them. This approach effectively always prefers the most likely character over all other alternatives at each step. The first result will thus be the same as the local optimum using a greedy approach, which becomes problematic for QAC where order is critical. By rebalancing the probability of the most likely suggestion with the average diversity weight given to other suggestions, we make sure probabilities stay uniform yet suggestions are diverse. We use a normalized Levenshtein distance to assess the diversity. 

\section{Experiments}
\label{sec:experiments}
\subsection{Dataset}
\label{sec:dataset}
The AOL query logs \cite{pass2006picture} are commonly used to evaluate the quality of QAC systems. We rely on a background dataset for the NN; training and validation datasets for lambdaMART integrations; and a test dataset for evaluations. Some adaptations are done to the AOL background dataset as in \cite{park2017neural}, such as removing the queries appearing less than 3 times or longer that 100 characters. For each query in the training, validation and test datasets, we use all possible prefixes starting after the first word as in \cite{shokouhi2013learning}. We use the sets from \cite{park2017neural} available online, enriched with user and time information provided in the original AOL dataset. In addition, \red{we evaluate the systems on a second real-world dataset from a production search engine in the biomedical domain, PubMed \cite{fiorini2017,lu2011pubmed,Mohan2018AFD}, that was created in the same manner.} The biomedical dataset consists of 8,490,317 queries. The sizes of training, validation and test sets are comparable to those used for the AOL dataset.

\subsection{Evaluation}
Systems are evaluated using the traditional Mean Reciprocal Rank (MRR) metric. This metric assesses the quality of suggestions by identifying the rank of the real query in the suggestions given one of its prefixes. We also tested PMRR as introduced in \cite{park2017neural} and observed the same trends in results as MRR, so we do not show them due to space limitation. Given the set of prefixes $P$ in the test dataset, MRR is defined as follows:
\begin{equation}
MRR = \frac{1}{|Q|}\sum_{r \in P}{\frac{1}{r_p}},
\end{equation}
where $r_p$ represent the rank of the match. Paired t-tests measure the significance of score variations among systems and are reported in the Results section. We also evaluate prediction time as this is an important parameter for building production systems. The prediction time is averaged over 10 runs on the test set, on the same hardware for all models. We do not evaluate throughput but rather compare the time required by all approaches to process one prefix.

\subsection{Systems and setups}
\label{sec:systems}
We implemented the method in \cite{park2017neural} and used their best-performing model as a baseline. We also compare our results to the standard \textbf{MPC} \cite{bar2011context}. For our method, we evaluate several incremental versions, starting with \textbf{NQAC} which follows the architecture detailed above but with the word embeddings and the one-hot encoding of characters only. We add the subscript \textbf{U} when the language model is enriched with user vectors and \textbf{T} when it integrates time features. We append \textbf{+D} to indicate the use of the diverse beam search to predict queries instead of a standard beam search. Finally, we also study the impact of adding MPC and LambdaMART (\textbf{+MPC}, \textbf{+$\lambda$MART}). 

\section{Results}
\label{sec:results}
A summary of the results is presented in Table \ref{tab:all_results}. \red{Interestingly, our simple NQAC model performs similarly to the state-of-the-art on this dataset, called Neural Query Language Model (NQLM), on all queries. It is significantly less good for seen queries (-5.6\%) and significantly better for unseen queries (+4.2\%).} Although GRUs have less expressive power than LSTMs, their smaller number of parameters to train allowed them to better converge than all LSTM models we tested, including that of \cite{park2017neural}. NQAC also benefits from a significantly better scalability (28\% faster than NQLM) and thus seems more appropriate for production systems. When we enrich the language model with user information, it becomes better for seen queries (+1.9\%) while being about as fast. Adding time sensitivity does not yield significant improvements on this dataset \red{overall, but improves significantly the performance for seen queries (+1.7\%)}. Relying on the diverse beam search significantly hurts the processing time (39\% longer) while not providing significantly better performance. Our integration of MPC differs from previous studies. We noticed that for Web search, MPC performs extremely well and is computationally cheap (0.24 seconds). On the other hand, all neural QAC systems are better for unseen queries but struggle to stay under a second of processing time. Since identifying if a query has been seen or not is done in constant time, we route the query either to MPC or to NQAC$_{UT}$ and we note the overall performance as NQAC$_{UT}$+MPC. This method provides a significant improvement over NQLM (+6.7\%) overall while being faster on average. Finally, appending NQAC$_{UT}$'s results to MPC's and reranking the list with LambdaMART provides the best results on this dataset, but at the expense of greater computational cost (+60\%).

\begin{table*}
  \caption{MRR results for all tested models on the AOL and biomedical datasets with their average prediction time in seconds.}
  \label{tab:all_results}
  \centering
  \resizebox{\textwidth}{!}{%
  \begin{tabular}{lcccccccc}
    \toprule
    &\multicolumn{4}{c}{AOL dataset}&\multicolumn{4}{c}{Biomedical dataset}\\
    \cmidrule(lr){2-5}
    \cmidrule(lr){6-9}
    \multirow{2}{*}{Model} &&MRR & &\multirow{2}{*}{Time} & & MRR & &\multirow{2}{*}{Time}\\
    \cmidrule(lr){2-4}
    \cmidrule(lr){6-8}
    &Seen&Unseen&All&&Seen& Unseen & All&\\
    \midrule
    MPC \cite{bar2011context} & \textbf{0.461} & 0.000 & 0.184&\textbf{0.24}& 0.165 & 0.000 & 0.046&\textbf{0.29}\\
    NQLM(L)+WE+MPC+$\lambda$MART \cite{park2017neural}&0.430&0.306&0.356&1.33&0.159&0.152&0.154&2.35\\
    \midrule
    \textbf{Our models in this paper}&&&&&&\\
    NQAC & 0.406 & 0.319 & 0.354 &0.94& 0.155 & 0.139 & 0.143&1.73\\
    NQAC$_{U}$ & 0.417 & 0.325 & 0.361 &0.98&\textbf{0.191}&0.161&0.169&1.77\\
    NQAC$_{UT}$ & 0.424 & 0.326 & 0.365 &0.95&0.101&\textbf{0.195}& 0.157&1.81\\
    NQAC$_{UT}$+D &0.427&0.326& 0.366 &1.32& 0.186 & 0.185&0.185&2.04\\
    NQAC$_{UT}$+MPC&\textbf{0.461}&0.326&0.380&0.68&0.165&\textbf{0.195}&\textbf{0.187}&1.20\\
    NQAC$_{UT}$+MPC+$\lambda$MART&0.459&\textbf{0.330}&\textbf{0.382}&1.09&0.154&0.179& 0.172&2.01\\
  \bottomrule
\end{tabular}
}
\end{table*}

While NQAC$_{UT}$+MPC appears clearly as the best compromise between performance and quality for the AOL dataset, the landscape changes drastically on the biomedical dataset and the quality drops significantly for all systems. This shows the potential difficulties associated with real-world systems, which particularly occur in specialized domains. In this case, the drop in performance is mostly due to the fact that biomedical queries are longer and it becomes more difficult for models to predict the entire query accurately only with the first keywords. \red{While the generated queries make sense and are relevant candidates, the chance for generative models to predict the exact target query diminishes as the target query is longer because of combinatorial explosion. This is even more true when the target queries are diverse as in specialized domains \cite{islamaj2009understanding,neveol2011semi}. For example, for the prefix ``breast cancer'', there are 1169 diverse suffixes in a single day of logs used for training. These include ``local recurrence'', ``nodular prognosis'', ``hormone receptor'', ``circulating cells'', ``family history'', ``chromosome 4p16'' or ``herceptin review'', to cite only a few. Hence, while the model predicts plausible queries, it is a lot more difficult to predict the one the user intended.} The target query length also has an impact on prediction time, as roughly twice the time is needed for Web searches. MPC is the exception, however, it performs poorly even on seen queries (0.165). This observation suggests that more elaborate models are specifically needed for specialized domains. On this dataset, NQAC does not perform as well as NQLM and it seems this time that the higher number of parameters in NQLM is more appropriate for the task. Still, user information helps significantly for seen queries (+23\%), probably because some users frequently check the same queries to keep up-to-date. Time sensitivity seems to help significantly unseen queries (+21\%) while significantly hurting the quality for seen queries (-47\%). Diversity is significantly helpful on this dataset (+19\%) and provides a balance in performance for both seen and unseen queries. NQAC$_{UT}$+MPC yields the best overall MRR score for this dataset, and LambdaMART is unable to learn how to re-rank the suggestions, thus decreasing the score.\\
From these results, we draw several conclusions. First, MPC performs very well on seen queries for Web searches and it should be used on them. For unseen queries, the NQAC$_{UT}$ model we propose achieves a sub-second state-of-the-art performance. Second, it is clear that the field of application will affect many of the decisions when designing a QAC system. On a specialized domain, the task is more challenging: fast approaches like MPC perform too poorly while more elaborate approaches do not meet production requirements. NQAC$_{U}$ performs best on seen queries, NQAC$_{UT}$ on unseen queries. Finally, NQAC$_{UT}$+D provides an equilibrium between the two at a greater computational cost. Its overall MRR is similar to that of NQAC$_{UT}$+MPC but it is less redundant (see Table \ref{tab:examples}). \red{Particularly, the system seems not to be limited anymore by the higher probability associated with shorter suggestions (e.g. ``www google'', a form of ``www google com''), thus bringing more diversity. This aspect can be more useful for specialized domains where the range of possible queries is broader. Finally, we found that a lot more data was needed for the biomedical domain than for general Web search. After about a million queries, NQAC suggests meaningful and plausible queries for both datasets. However, for the biomedical dataset, the loss needs more epochs to stabilize than for the AOL dataset, mainly due to the combinatorial explosion mentioned above.}

\begin{table}
  \caption{Comparison of the 10 top query candidates from the baselines and our approach for the prefix ``www''.}
  \label{tab:examples}
  \centering
  \resizebox{\columnwidth}{!}{%
  \begin{tabular}{lll}
    \toprule
    MPC & \cite{park2017neural} & NQAC+D\\
    \midrule
    www google com & www google com & www google com\\
    www yahoo com & www yahoo com & www myspace com\\
    www myspace com & www myspace com & www mapquest com\\
    www google & www google & www yahoo com\\
    www ebay com & www hotmail com & www hotmail com\\
    www hotmail com & www my & www bankofamerica com \\
    www mapquest com & www myspace com & www chase com\\
    www myspace & www mapquest com & www disneychannel com\\
    www msn com & www yahoo & www myspace \\
    www bankofamerica com & www disney channel com & www disney channel com\\
  \bottomrule
\end{tabular}
}
\end{table}

\section{Conclusions and future work}
To the best of our knowledge, we proposed the first neural language model that integrates user information and time sensitivity for query auto completion with a focus on scalability for real-world systems. Personalization is provided through pretrained user vectors based on their past queries. By incorporating this information and by adapting the architecture, we were able to achieve state-of-the-art performance in neural query auto completion without relying on re-ranking, making this approach significantly more scalable in practice. We studied multiple variants, their benefits and drawbacks for various use cases. We also demonstrate the utility of this method for specialized domains such as biomedicine, where the query diversity and vocabulary are broader and MPC fails to provide the same performance as in Web search. We also found that user information and diversity improve the performance significantly more than for Web search engines. To allow readers to easily reproduce, evaluate and improve our models, we provide all the code on a public repository.\\
The handling of time-sensitivity may benefit from a more elaborate integration, for example session-based rather than absolute time. Also, we evaluated our approaches on a general search setup for both datasets, while searches in the biomedical domain commonly contain fields (i.e. authors, title, abstract, etc.) which adds to the difficulty. The choice of a diversity metric is also important and could be faster or more efficient (e.g., using word embeddings to diversify the semantics of the suggestions). These limitations warrant further work and we leave them as perspectives.

\section*{Acknowledgement}
\red{This research was supported by the Intramural Research Program of the NIH, National Library of Medicine.}

\bibliography{naaclhlt2018}

\begin{thebibliography}{26}
\expandafter\ifx\csname natexlab\endcsname\relax\def\natexlab#1{#1}\fi

\bibitem[{Bar-Yossef and Kraus(2011)}]{bar2011context}
Ziv Bar-Yossef and Naama Kraus. 2011.
\newblock Context-sensitive query auto-completion.
\newblock In \emph{Proceedings of the 20th international conference on World
  wide web}, pages 107--116. ACM.

\bibitem[{Burges(2010)}]{burges2010ranknet}
Christopher~JC Burges. 2010.
\newblock From ranknet to lambdarank to lambdamart: An overview.
\newblock \emph{Learning}, 11(23-581):81.

\bibitem[{Cai et~al.(2016)Cai, De~Rijke et~al.}]{cai2016survey}
Fei Cai, Maarten De~Rijke, et~al. 2016.
\newblock A survey of query auto completion in information retrieval.
\newblock \emph{Foundations and Trends{\textregistered} in Information
  Retrieval}, 10(4):273--363.

\bibitem[{Cai et~al.(2014)Cai, Liang, and De~Rijke}]{cai2014time}
Fei Cai, Shangsong Liang, and Maarten De~Rijke. 2014.
\newblock Time-sensitive personalized query auto-completion.
\newblock In \emph{Proceedings of the 23rd ACM international conference on
  conference on information and knowledge management}, pages 1599--1608. ACM.

\bibitem[{Cho et~al.(2014)Cho, Van~Merri{\"e}nboer, Bahdanau, and
  Bengio}]{cho2014properties}
Kyunghyun Cho, Bart Van~Merri{\"e}nboer, Dzmitry Bahdanau, and Yoshua Bengio.
  2014.
\newblock On the properties of neural machine translation: Encoder-decoder
  approaches.
\newblock \emph{arXiv preprint arXiv:1409.1259}.

\bibitem[{Fiorini et~al.(2017)Fiorini, Lipman, and Lu}]{fiorini2017}
Nicolas Fiorini, David~J Lipman, and Zhiyong Lu. 2017.
\newblock Cutting edge: Towards pubmed 2.0.
\newblock \emph{eLife}, 6:e28801.

\bibitem[{Islamaj~Dogan et~al.(2009)Islamaj~Dogan, Murray, N{\'e}v{\'e}ol, and
  Lu}]{islamaj2009understanding}
Rezarta Islamaj~Dogan, G~Craig Murray, Aur{\'e}lie N{\'e}v{\'e}ol, and Zhiyong
  Lu. 2009.
\newblock Understanding pubmed{\textregistered} user search behavior through
  log analysis.
\newblock \emph{Database}, 2009.

\bibitem[{Jozefowicz et~al.(2015)Jozefowicz, Zaremba, and
  Sutskever}]{jozefowicz2015empirical}
Rafal Jozefowicz, Wojciech Zaremba, and Ilya Sutskever. 2015.
\newblock An empirical exploration of recurrent network architectures.
\newblock In \emph{International Conference on Machine Learning}, pages
  2342--2350.

\bibitem[{Kim et~al.(2016)Kim, Jernite, Sontag, and Rush}]{kim2016character}
Yoon Kim, Yacine Jernite, David Sontag, and Alexander~M Rush. 2016.
\newblock Character-aware neural language models.
\newblock In \emph{AAAI}, pages 2741--2749.

\bibitem[{Koutrika and Ioannidis(2005)}]{koutrika2005unified}
Georgia Koutrika and Yannis Ioannidis. 2005.
\newblock A unified user profile framework for query disambiguation and
  personalization.
\newblock In \emph{Proceedings of workshop on new technologies for personalized
  information access}, pages 44--53.

\bibitem[{Lankinen et~al.(2016)Lankinen, Heikinheimo, Takala, Raiko, and
  Karhunen}]{lankinen2016character}
Matti Lankinen, Hannes Heikinheimo, Pyry Takala, Tapani Raiko, and Juha
  Karhunen. 2016.
\newblock A character-word compositional neural language model for finnish.
\newblock \emph{arXiv preprint arXiv:1612.03266}.

\bibitem[{Le and Mikolov(2014)}]{le2014distributed}
Quoc Le and Tomas Mikolov. 2014.
\newblock Distributed representations of sentences and documents.
\newblock In \emph{International Conference on Machine Learning}, pages
  1188--1196.

\bibitem[{Lu(2011)}]{lu2011pubmed}
Zhiyong Lu. 2011.
\newblock Pubmed and beyond: a survey of web tools for searching biomedical
  literature.
\newblock \emph{Database}, 2011.

\bibitem[{Lu et~al.(2009)Lu, Wilbur, McEntyre, Iskhakov, and
  Szilagyi}]{lu2009finding}
Zhiyong Lu, W~John Wilbur, Johanna~R McEntyre, Alexey Iskhakov, and Lee
  Szilagyi. 2009.
\newblock Finding query suggestions for pubmed.
\newblock In \emph{AMIA Annual Symposium Proceedings}, volume 2009, page 396.
  American Medical Informatics Association.

\bibitem[{Margaris et~al.(2018)Margaris, Vassilakis, and
  Georgiadis}]{margaris2018query}
Dionisis Margaris, Costas Vassilakis, and Panagiotis Georgiadis. 2018.
\newblock Query personalization using social network information and
  collaborative filtering techniques.
\newblock \emph{Future Generation Computer Systems}, 78:440--450.

\bibitem[{Mikolov et~al.(2013)Mikolov, Sutskever, Chen, Corrado, and
  Dean}]{mikolov2013distributed}
Tomas Mikolov, Ilya Sutskever, Kai Chen, Greg~S Corrado, and Jeff Dean. 2013.
\newblock Distributed representations of words and phrases and their
  compositionality.
\newblock In \emph{Advances in neural information processing systems}, pages
  3111--3119.

\bibitem[{Mitra and Craswell(2015)}]{mitra2015query}
Bhaskar Mitra and Nick Craswell. 2015.
\newblock Query auto-completion for rare prefixes.
\newblock In \emph{Proceedings of the 24th ACM international on conference on
  information and knowledge management}, pages 1755--1758. ACM.

\bibitem[{Mohan et~al.(2018)Mohan, Fiorini, Kim, and Lu}]{Mohan2018AFD}
Sunil Mohan, Nicolas Fiorini, Sun Kim, and Zhiyong Lu. 2018.
\newblock A fast deep learning model for textual relevance in biomedical
  information retrieval.
\newblock \emph{CoRR}, abs/1802.10078.

\bibitem[{N{\'e}v{\'e}ol et~al.(2011)N{\'e}v{\'e}ol, Do{\u{g}}an, and
  Lu}]{neveol2011semi}
Aur{\'e}lie N{\'e}v{\'e}ol, Rezarta~Islamaj Do{\u{g}}an, and Zhiyong Lu. 2011.
\newblock Semi-automatic semantic annotation of pubmed queries: a study on
  quality, efficiency, satisfaction.
\newblock \emph{Journal of biomedical informatics}, 44(2):310--318.

\bibitem[{Park and Chiba(2017)}]{park2017neural}
Dae~Hoon Park and Rikio Chiba. 2017.
\newblock A neural language model for query auto-completion.
\newblock In \emph{Proceedings of the 40th International ACM SIGIR Conference
  on Research and Development in Information Retrieval}, pages 1189--1192. ACM.

\bibitem[{Pass et~al.(2006)Pass, Chowdhury, and Torgeson}]{pass2006picture}
Greg Pass, Abdur Chowdhury, and Cayley Torgeson. 2006.
\newblock A picture of search.
\newblock In \emph{InfoScale}, volume 152, page~1.

\bibitem[{Shokouhi(2013)}]{shokouhi2013learning}
Milad Shokouhi. 2013.
\newblock Learning to personalize query auto-completion.
\newblock In \emph{Proceedings of the 36th international ACM SIGIR conference
  on Research and development in information retrieval}, pages 103--112. ACM.

\bibitem[{Shokouhi and Radinsky(2012)}]{shokouhi2012time}
Milad Shokouhi and Kira Radinsky. 2012.
\newblock Time-sensitive query auto-completion.
\newblock In \emph{Proceedings of the 35th international ACM SIGIR conference
  on Research and development in information retrieval}, pages 601--610. ACM.

\bibitem[{Sutskever et~al.(2011)Sutskever, Martens, and
  Hinton}]{sutskever2011generating}
Ilya Sutskever, James Martens, and Geoffrey~E Hinton. 2011.
\newblock Generating text with recurrent neural networks.
\newblock In \emph{Proceedings of the 28th International Conference on Machine
  Learning (ICML-11)}, pages 1017--1024.

\bibitem[{Swiffin et~al.(1987)Swiffin, Arnott, Pickering, and
  Newell}]{swiffin1987adaptive}
Andrew Swiffin, John Arnott, J~Adrian Pickering, and Alan Newell. 1987.
\newblock Adaptive and predictive techniques in a communication prosthesis.
\newblock \emph{Augmentative and Alternative Communication}, 3(4):181--191.

\bibitem[{Vijayakumar et~al.(2016)Vijayakumar, Cogswell, Selvaraju, Sun, Lee,
  Crandall, and Batra}]{vijayakumar2016diverse}
Ashwin~K Vijayakumar, Michael Cogswell, Ramprasath~R Selvaraju, Qing Sun,
  Stefan Lee, David Crandall, and Dhruv Batra. 2016.
\newblock Diverse beam search: Decoding diverse solutions from neural sequence
  models.
\newblock \emph{arXiv preprint arXiv:1610.02424}.

\end{thebibliography}
\bibliographystyle{acl_natbib}

\appendix

\end{document}